\documentclass[10pt]{article}
\usepackage{spconf}
\usepackage{amsmath,graphicx,hyperref}

\usepackage{amssymb,amsfonts}
\usepackage{algorithmic}
\usepackage{textcomp}
\usepackage{xcolor}
\usepackage{bm}
\usepackage{url}
\usepackage{booktabs}
\usepackage{bm}
\usepackage{cite}
\usepackage{eso-pic}

\DeclareMathOperator*{\minimize}{minimize}

\title{WEEP: A Differentiable Nonconvex Sparse Regularizer via Weakly-Convex Envelope}
\name{
    Takanobu Furuhashi\textsuperscript{1}, Hidekata Hontani\textsuperscript{1}, Qibin Zhao\textsuperscript{2}, Tatsuya Yokota\textsuperscript{1,2}
}
\address{
    \textsuperscript{1}Nagoya Institute of Technology, Aichi, Japan \\
    \textsuperscript{2}RIKEN Center for Advanced Intelligence Project, Tokyo, Japan
    \thanks{This work was partially supported by the Japan Society for the Promotion of Science (JSPS) KAKENHI under Grant 23K28109.}
}

\begin{document}
%
\maketitle
\AddToShipoutPictureBG*{%
    \AtPageUpperLeft{%
        \makebox[\paperwidth]{%
            \parbox[t][2.5cm][c]{0.7\paperwidth}{%
                \centering  
                \footnotesize
                \textcopyright 2026 IEEE. Personal use of this material is permitted. Permission from IEEE must be obtained for all other uses, in any current or future media, including reprinting/republishing this material for advertising or promotional purposes, creating new collective works, for resale or redistribution to servers or lists, or reuse of any copyrighted component of this work in other works.
                
                Accepted for publication in 2026 IEEE International Conference on Acoustics, Speech and Signal Processing (ICASSP 2026).
            }%
        }%
    }%
}
\begin{abstract}
    Sparse regularization is fundamental in signal processing and feature extraction but often relies on non-differentiable penalties, conflicting with gradient-based optimizers. We propose WEEP (Weakly-convex Envelope of Piecewise Penalty), a novel differentiable regularizer derived from the weakly-convex envelope framework. WEEP provides tunable, unbiased sparsity and a simple closed-form proximal operator, while maintaining full differentiability and L-smoothness, ensuring compatibility with both gradient-based and proximal algorithms. This resolves the tradeoff between statistical performance and computational tractability. We demonstrate superior performance compared to established convex and non-convex sparse regularizers on challenging compressive sensing and image denoising tasks.
\end{abstract}
\section{Introduction}
\label{sec:introduction}

\begin{figure}[t]
    \centering
    \includegraphics[width=0.9\linewidth]{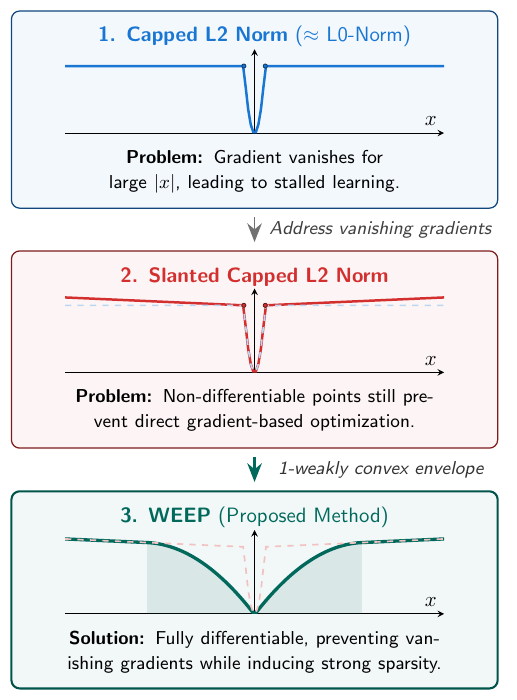}
    \caption{WEEP design flow: Evolution from Capped L2 Norm to the proposed WEEP method ($a = 100$, $b = 0.025$).
    }
    \label{fig:weep_design_flow}
\end{figure}

\begin{table*}[t]
    \centering
    \caption{
        Comparison of WEEP with major regularization paradigms. WEEP is the only method that simultaneously satisfies desirable properties for sparse regularization, making it a versatile choice for signal processing and machine learning tasks.
    }
    \vspace*{5pt}
    \label{tab:regularizer_comparison}
    \resizebox{0.95\textwidth}{!}{%
        \begin{tabular}{l|c|cc|ccc}
            \toprule
                                         & \textbf{Proposed}           & \multicolumn{2}{c|}{\textbf{Non-Differentiable Penalties}} & \multicolumn{3}{c}{\textbf{Differentiable Alternatives}}                                                                                  \\
            \cmidrule(lr){2-2} \cmidrule(lr){3-4} \cmidrule(lr){5-7}
            \textbf{Property}            & \textbf{WEEP}               & \textbf{L1 (Lasso)}                                        & \textbf{MCP / SCAD}                                      & \textbf{L2 / Huber}      & \textbf{Smooth L0}                   & \textbf{DWF} \\
            \midrule
            \multicolumn{7}{l}{\textit{\textbf{Statistical Performance}}}                                                                                                                                                                                                       \\
            \quad Sparsity Power         & \textbf{Low $\sim$ High}    & Medium                                                     & High                                                     & Low                      & Low $\sim$ High                      & High         \\
            \quad Estimation Bias        & \textbf{Unbiased} ($b = 0$) & Biased                                                     & Unbiased                                                 & Biased                   & \multicolumn{2}{|c}{Nearly Unbiased}                \\
            \midrule
            \multicolumn{7}{l}{\textit{\textbf{Optimization-Friendliness}}}                                                                                                                                                                                                     \\
            \quad Vanishing Gradients    & \textbf{No} ($b>0$)         & No                                                         & Yes                                                      & No                       & Yes                                  & No           \\
            \quad Closed-form Proximal   & \textbf{Yes}                & \multicolumn{2}{c|}{Yes}                                   & Yes                                                      & N/A                      & N/A                                                 \\
            \quad Differentiability      & \textbf{Full}               & \multicolumn{2}{c|}{No (at origin)}                        & \multicolumn{2}{c|}{Full}                                & Full (w.r.t. new params)                                                       \\
            \quad Weak Convexity         & \textbf{Yes}                & \multicolumn{2}{|c|}{Yes}                                  & \multicolumn{2}{|c|}{Yes}                                & Dependent on Objective                                                         \\
            \midrule
            \multicolumn{7}{l}{\textit{\textbf{Practical Applicability}}}                                                                                                                                                                                                       \\
            \quad Feature Sparsification & \textbf{Yes}                & \multicolumn{2}{c|}{Yes}                                   & \multicolumn{2}{c|}{Yes}                                 & Dependent on solver                                                            \\
            \quad Data Fidelity Usage    & \textbf{Yes}                & \multicolumn{2}{c|}{Yes}                                   & \multicolumn{2}{c|}{Yes}                                 & No (Regularizer-only)                                                          \\
            \bottomrule
        \end{tabular}%
    }
    \vspace*{-10pt}
\end{table*}

Sparse regularization has become an indispensable tool in modern signal processing, providing a powerful paradigm for solving ill-posed inverse problems \cite{wenSurveyNonconvexRegularizationBased2018,bruntonDatadrivenScienceEngineering2019,crespomarquesReviewSparseRecovery2019,tianComprehensiveSurveyRegularization2022}. Its applications span from medical imaging \cite{yeCompressedSensingMRI2019,luIterativeReconstructionLowdose2023} to seismic data analysis \cite{yuInterpolationDenoisingHighdimensional2015,chenDoublesparsityDictionarySeismic2016,zhaoSignalPreservingErraticNoise2018} and wireless communications \cite{gaoSpatiallyCommonSparsity2015,heCascadedChannelEstimation2020}. The core principle exploits the fact that many natural signals exhibit sparsity in appropriate bases \cite{yeCompressedSensingMRI2019,bruntonDatadrivenScienceEngineering2019,9729560}.

The L1-norm \cite{10.2307/2346178} has been the common choice due to its convexity, but suffers from estimation bias that shrinks large, significant coefficients towards zero, degrading signal fidelity \cite{selesnickSparseRegularizationConvex2017,wenSurveyNonconvexRegularizationBased2018,sasakiSparseRegularizationBased2024}. Non-convex penalties like Smoothly Clipped Absolute Deviation (SCAD) \cite{fanVariableSelectionNonconcave2001} and Minimax Concave Penalty (MCP) \cite{zhangNearlyUnbiasedVariable2010} reduce bias by saturating penalties for large coefficients \cite{wenSurveyNonconvexRegularizationBased2018,tianComprehensiveSurveyRegularization2022,sasakiSparseRegularizationBased2024}. However, their non-differentiability and vanishing gradients make them incompatible with efficient gradient-based optimizers like Stochastic Gradient Descent (SGD), Adam, Limited-memory BFGS (L-BFGS) and bilevel optimizers \cite{liuLimitedMemoryBFGS1989,kingmaAdamMethodStochastic2017,sunOptimizationDeepLearning2020,zhangIntroductionBilevelOptimization2024}.
This typically requires proximal methods like Alternating Direction Method of Multipliers (ADMM) \cite{hanSurveyRecentDevelopments2022}, which are often computationally expensive \cite{dixitOnlineLearningInexact2019}.
Moreover, the overly aggressive sparsity-inducing behavior of SCAD and MCP often leads to degraded performance in applications like image denoising \cite{sasakiSparseRegularizationBased2024}.

We resolve these limitations by proposing \textbf{WEEP (Weakly-Convex Envelope of Piecewise Penalty)}. WEEP uses the 1-weakly convex envelope \cite{yukawaContinuousRelaxationDiscontinuous2025} of a non-convex base penalty designed to approximate the L0-norm and be differentiable at the origin (see the middle plot in Fig.~\ref{fig:weep_design_flow}), yielding a regularizer that retains the benefits of the L0-norm while being fully differentiable everywhere, L-smooth, and weakly-convex. This makes WEEP a drop-in replacement for gradient-based optimizers like SGD, Adam and L-BFGS.

Our main contributions are: (1) The WEEP regularizer offering unbiased and tunable sparsity, L-smoothness, and a closed-form proximal operator \cite[Chapter 24]{bauschkeConvexAnalysisMonotone2017} as shown in Table~\ref{tab:regularizer_comparison}. (2) Comprehensive validation on compressive sensing and real-world noisy image denoising tasks, demonstrating superior performance and efficiency.

\section{Preliminaries}

\subsection{Sparse Regularization in Signal Processing}
The typical sparse regularization problem has the form:
\begin{equation}
    \minimize_{\bm{x}\in\mathbb{R}^n} f(\bm{x}) + \lambda \sum_{i=1}^d \phi((\bm{L}\bm{x})_i)
\end{equation}
where $f(\bm{x})$ is a data fidelity term (typically the squared error), $\phi(\cdot)$ promotes sparsity, $\bm{L}$ is the linear operator such as the Fourier transform or the finite difference, and $\lambda \geq 0$ controls regularization strength. Non-convex penalties like SCAD \cite{fanVariableSelectionNonconcave2001}, MCP \cite{zhangNearlyUnbiasedVariable2010}, Capped L2 Norm \cite{strekalovskiyRealTimeMinimizationPiecewise2014} and Smooth L0 \cite{mohimaniFastApproachOvercomplete2009} reduce bias by saturating for inputs with large and significant absolute values, but their vanishing gradients limit gradient-based optimization, requiring proximal solvers \cite{beckFirstOrderMethodsOptimization2017,parikhProximalAlgorithms2014}.
Deep Weight Factorization (DWF) \cite{kolbDeepWeightFactorization2024} provides a differentiable and non-vanishing gradient alternative, but it requires more numerical parameters and cannot be readily used as a generic penalty term. Moreover, its applicability is generally limited to the case where $\bm{L} = \bm{I}$ (identity).

\subsection{Mathematical Requirements for Optimization}
For differentiable and L-smooth functions (L-Lipschitz continuous gradient), gradient descent converges via monotonic descent \cite{nesterovIntroductoryLecturesConvex2004}. Non-smooth problems require proximal methods like ADMM \cite{hanSurveyRecentDevelopments2022}, whose efficiency depends on closed-form proximal operators. Weak convexity plays a crucial role for theoretical convergence guarantees in non-convex optimization, as it ensures that proximal methods avoid problematic behaviors mainly by guaranteeing the proximal operator is single-valued \cite{hanSurveyRecentDevelopments2022,yukawaMonotoneLipschitzGradientDenoiser2025a,yukawaContinuousRelaxationDiscontinuous2025}. WEEP provides L-smoothness, weak convexity, and a closed-form proximal operator for both gradient-based and proximal optimization.

A function $f$ is $\mu$-weakly convex if $f(x) + \frac{\mu}{2}x^2$ is convex. The 1-weakly convex envelope $f^{1\text{-wce}}$ is the tightest 1-weakly convex function below $f$ \cite[Definition 3]{yukawaContinuousRelaxationDiscontinuous2025}. This construction systematically smooths non-differentiable kinks and fills non-convex regions while preserving essential global structure. A notable example involves the 1-weakly convex envelope of the L0-norm, which yields the MCP \cite{yukawaContinuousRelaxationDiscontinuous2025}. For certain piecewise functions, this envelope can be constructed analytically, making it practical for designing optimization-friendly regularizers that retain benefits of non-convex counterparts.

\section{Proposed Method: WEEP}
Our approach consists of two steps: (1) designing a flexible, non-convex base penalty with desirable statistical properties, and (2) applying the 1-weakly convex envelope to transform this penalty into a smooth, optimization-friendly regularizer.

\begin{figure}[t]
    \centering
    \includegraphics[width=0.8\columnwidth]{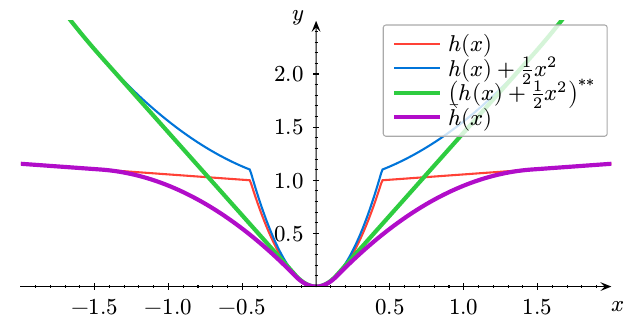}
    \caption{Construction of WEEP with $a = 10$, $b = 0.1$.}
    \label{fig:weep_construction}
\end{figure}

First, we define the Capped L2 Norm with a linear penalty for large inputs as our base penalty $h(x)$:
\begin{equation}
    h(x) = \begin{cases}
        \frac{a}{2}x^2 & \text{if } |x| \le \sqrt{2/a} \\
        b|x| + c       & \text{if } |x| > \sqrt{2/a}
    \end{cases}
\end{equation}
where $a > 0$ controls sparsity-inducing curvature at the origin, $b \ge 0$ sets a linear penalty for large inputs preventing vanishing gradients, and $c = 1 - b\sqrt{2/a}$ ensures continuity. Note that $h(x)$ approaches the L0-norm as $a \to \infty$ and $b \to 0$. However, this penalty remains highly non-convex and non-differentiable at transition points $|x| = \sqrt{2/a}$.

The proposed WEEP $\bar{h}(x)$ is constructed as the 1-weakly convex envelope of $h(x)$:
\begin{equation}
    \bar{h}(x) = \left(h(x) + \frac{1}{2}x^2\right)^{**} - \frac{1}{2}x^2
\end{equation}
where $(\cdot)^{**}$ denotes the biconjugate \cite[Definition 13.1]{bauschkeConvexAnalysisMonotone2017}. Its epigraph is the convex hull of that of $h(x) + \frac{1}{2}x^2$.
As shown in Fig.~\ref{fig:weep_construction}, this process fills the non-convex regions of $h(x) + \frac{1}{2}x^2$ with its convex envelope, resulting in the differentiable and weakly convex WEEP penalty $\bar{h}(x)$.
This yields:
\begin{equation}
    \bar{h}(x) = \begin{cases}
        \frac{a}{2}x^2            & |x| \le x_1       \\
        m|x| + d - \frac{1}{2}x^2 & x_1 < |x| \le x_2 \\
        b|x| + c                  & |x| > x_2
    \end{cases}
\end{equation}
where $x_1 = {m}/{(a+1)}$, $x_2 = m - b$, $d = -(m-b)^2 /2+ c$, and $m = a^{-1}\{b(a+1) + \sqrt{a+1}\cdot|b - \sqrt{2a}|\}$.

\subsection{Derivation of the WEEP Parameters}
\label{app:derivation_weep_parameters}

The parameters $x_1, x_2, m, d$ of WEEP are determined by the common tangent to the parabolic segments of $k(x) = h(x) + \frac{1}{2}x^2$. This construction is fundamentally related to the biconjugate of the function $k(x)$, which fills the non-convex regions of the epigraph by taking the convex hull (see Fig.~\ref{fig:weep_construction}). For $x > 0$, let $k_{\mathrm{in}}(x) = \frac{a+1}{2}x^2$ and $k_{\mathrm{out}}(x) = \frac{1}{2}x^2 + bx + c$, with common tangent line $L(x) = mx+d$.
The tangency conditions are: (1) Value equality: $k_{\mathrm{in}}(x_1) = L(x_1)$ and $k_{\mathrm{out}}(x_2) = L(x_2)$. (2) Derivative equality: $k_{\mathrm{in}}'(x_1)\!=\!L'(x_1)$ and $k_{\mathrm{out}}'(x_2)\!=\!L'(x_2)$.

From the derivative conditions, we immediately get expressions for $x_1 = {m}/{(a+1)}$ and $x_2 = m - b$.
We also obtain two expressions for $d$ from the value conditions:
\begin{align}
    d & = (a+1)x_1^2/2 - mx_1 = -(a+1)^{-1}m^2/2,   \\
    d & = x_2^2/2 + (b-m)x_2 + c = -(m-b)^2/2 + c.
\end{align}
Equating these and using $c = 1-b\sqrt{{2}/{a}}$ yields the equation:
\begin{equation}
    am^2 - 2b(a+1)m + (b^2-2c)(a+1) = 0.
\end{equation}
This can be solved using the quadratic formula:
\begin{equation}
    m = a^{-1}\left\{b(a+1) \pm \sqrt{a+1}\cdot|b - \sqrt{2a}|\right\}.
\end{equation}
We choose the positive root to ensure $m > 0$ and $x_1 \leq x_2$.

\subsection{Key Properties}
WEEP has several desirable properties for optimization. (1) \textbf{Full Differentiability}: Enables seamless integration with gradient-based optimizers. (2) \textbf{L-Smoothness}: Lipschitz continuous gradient with constant $L = \max(1, a)$, ensuring stable convergence via monotonic descent \cite{nesterovIntroductoryLecturesConvex2004}. (3) \textbf{Weak Convexity}: Guaranteed 1-weak convexity via the envelope construction, independent of hyperparameters \cite{hanSurveyRecentDevelopments2022,yukawaMonotoneLipschitzGradientDenoiser2025a,yukawaContinuousRelaxationDiscontinuous2025}. (4) \textbf{Closed-Form Proximal Operator}: Simple closed-form solution avoiding iterative solvers on proximal optimization. (5) \textbf{Non-Vanishing Gradients}: For $b>0$, prevents optimization stalls.
Table~\ref{tab:regularizer_comparison} provides a comprehensive comparison between WEEP and major regularizations, highlighting WEEP's unique combination of desirable properties.
The Lipschitz continuous gradient is given by
\begin{equation}
    \bar{h}'(x) = \begin{cases}
        ax                      & |x| \le x_1       \\
        m\cdot\text{sgn}(x) - x & x_1 < |x| \le x_2 \\
        b\cdot\text{sgn}(x)     & |x| > x_2
    \end{cases}
\end{equation}
and the closed-form proximal operator is given by:
\begin{equation}
    \mathrm{prox}_{\lambda \bar{h}}(z) = \begin{cases}
        \frac{z}{1+\lambda a}                               & \text{if } |z| \le z_1       \\
        \frac{z - \lambda m \cdot \text{sgn}(z)}{1-\lambda} & \text{if } z_1 < |z| \le z_2 \\
        z - \lambda b \cdot \text{sgn}(z)                   & \text{if } |z| > z_2
    \end{cases}
    \label{eq:proximal_operator_weep}
\end{equation}
where $z_1 = (1+\lambda a)x_1$ and $z_2 = (1-\lambda)x_2 + \lambda m$.
By setting a large $a$ and a small $b$, the proximal operator approximates hard-thresholding: it preserves values above a certain threshold with minimal shrinkage while shrinking smaller values towards zero.
Additionally, the tunable sparsity strength (via parameter $a$) allows for a better trade-off between preserving fine textures and removing noise. Its behavior is similar to Smooth L0 as discussed in \cite[Theorem 3]{mohimaniFastApproachOvercomplete2009}. This is a key advantage in signal and image recovery (see Section \ref{sec:experiments}).

\section{Experiments}
\label{sec:experiments}

\begin{figure}[t]
    \centering
    \includegraphics[width=0.975\columnwidth]{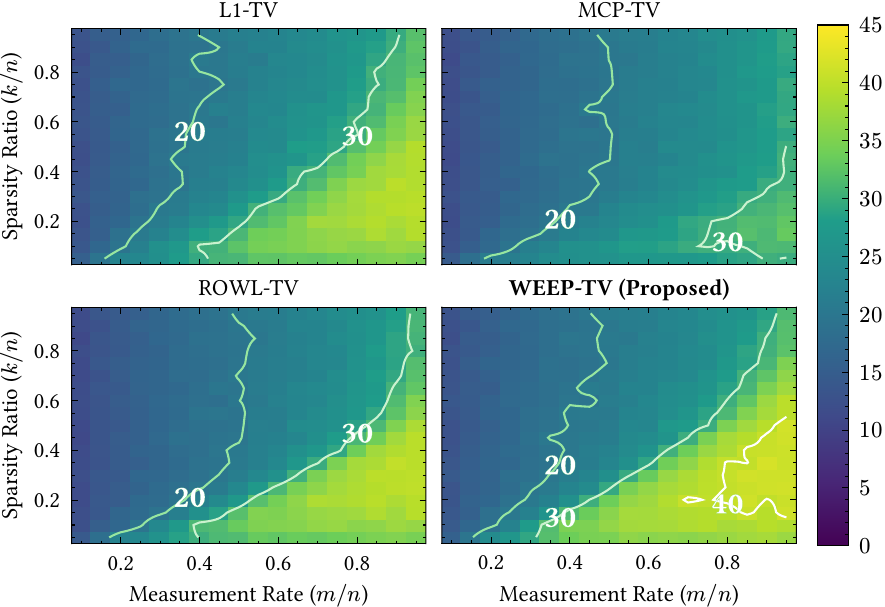}
    \caption{Phase transition analysis of recovery SNR (dB) for varying measurement rates ($m/n$) and sparsity ratios ($k/n$).}
    \label{fig:phase_transition}
\end{figure}

\begin{figure}[t]
    \centering
    \includegraphics[width=0.95\columnwidth]{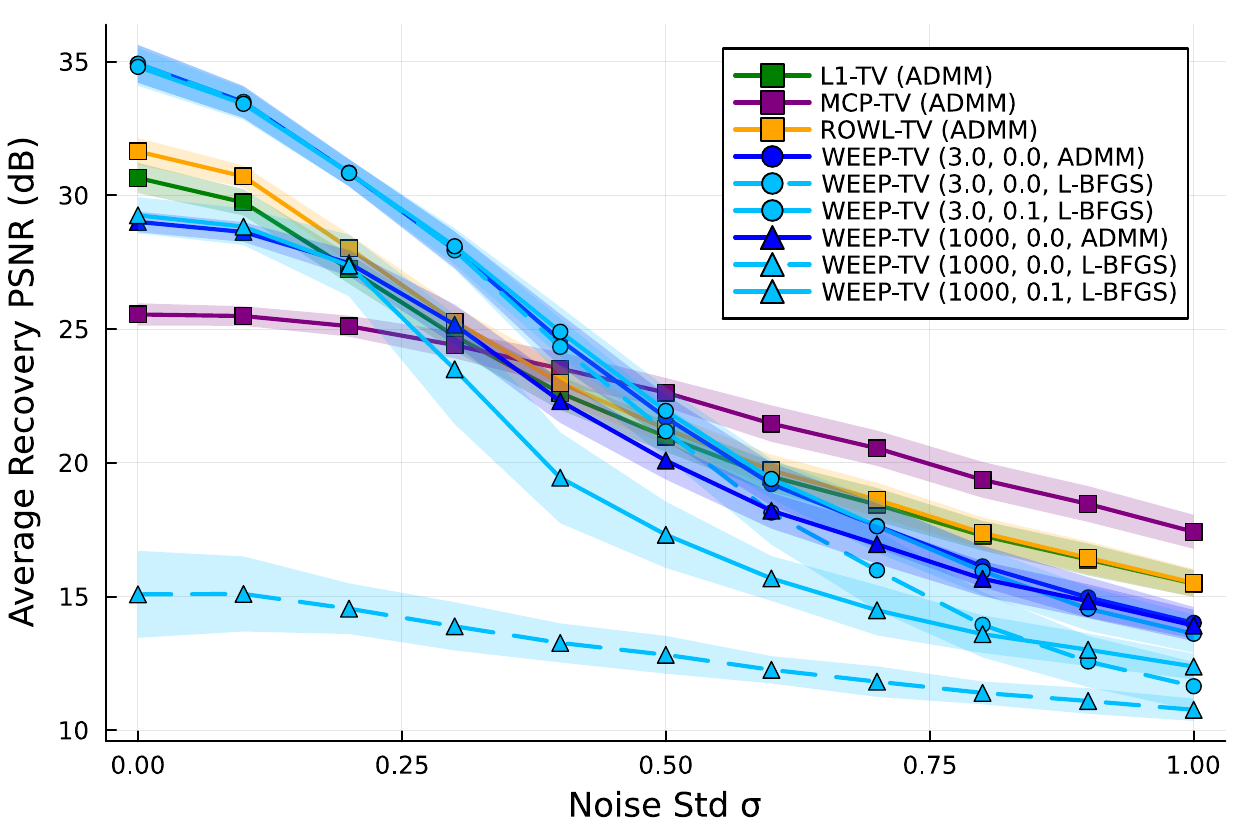}
    \caption{Average recovery PSNR as a function of the noise level over 100 trials with $m/n = 0.5$. Shaded ribbons represent $\pm$ one standard deviation.}
    \label{fig:cs_psnr_vs_noise}
\end{figure}

\begin{figure}[t]
    \centering
    \includegraphics[width=0.95\linewidth]{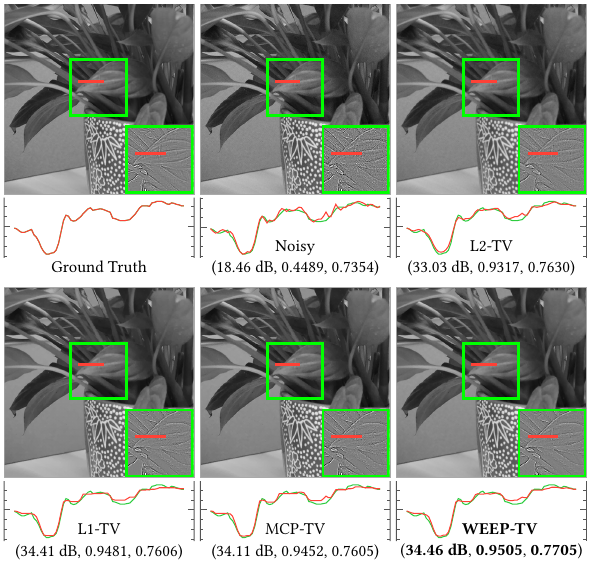}
    \caption{Real-world denoising results for the 'plant' image with performance evaluated in (PSNR↑, SSIM↑, FSIM↑).}
    \label{fig:image_denoising_results_with_insets}
\end{figure}
We conduct experiments on compressive sensing (CS) and real-world image denoising tasks to demonstrate WEEP's superior performance compared to established baselines.

\subsection{1D Compressive Sensing with TV Regularization}

We evaluate WEEP on a 1D Total Variation (TV) regularized CS task, where a signal $\bm{x} \in \mathbb{R}^n$ is recovered from a few noisy measurements $\bm{y} = \bm{A}\bm{x} + \bm{\varepsilon} \in \mathbb{R}^m$ ($m < n$) by solving: 
\begin{equation}
    \minimize_{\bm{x} \in \mathbb{R}^n} \frac{1}{2}\|\bm{y} - \bm{A}\bm{x}\|_2^2 + \lambda \sum_i \phi((\bm{D}\bm{x})_i),
\end{equation}
where $\bm A$ is a measurement matrix, $\bm \varepsilon \sim \mathcal{N}(0, \sigma^2\bm I)$, $\phi$ is a sparse regularizer and $\bm{D}\bm{x}$ denotes the discrete gradient of $\bm{x}$.

\textbf{Experimental Setup:} We generated synthetic signals ($n=512$) combining piecewise constant blocks ($k$ jumps) with fine-scale textures (triangle wave) to evaluate the preservation of both sharp edges and subtle details.
Compressed measurements were obtained using a Gaussian random matrix $\bm{A}$ and additive Gaussian noise $\bm \varepsilon$ with $\sigma = 0.05$.
We compared WEEP-TV against \{L1, MCP, ROWL\}-TV \cite{sasakiSparseRegularizationBased2024}.

\textbf{Phase Transition Analysis:}
We conducted a phase transition analysis to evaluate robustness across varying measurement rates ($m/n \in [0.1, 0.95]$) and sparsity ratios ($k/n \in [0.05, 0.95]$).
For each grid point, we averaged the recovery SNR over 50 trials, tuning hyperparameters (WEEP: $a \in \{1, 3, 10\}$, $b = 0$ for unbiasedness) to maximize SNR for each method.
Fig.~\ref{fig:phase_transition} shows the results.
WEEP-TV exhibits a significantly higher SNR compared to \{L1, MCP, ROWL\}-TV, particularly in challenging regimes with low measurement rates.
This indicates that WEEP's tunable non-convexity ($a$) allows for more accurate recovery of complex and fine-textured signals where other methods fail.

\textbf{Noise Robustness:}
We also evaluated robustness to noise by varying the noise level $\sigma$ (Fig.~\ref{fig:cs_psnr_vs_noise}) with fixed $k=12$ and $m/n = 0.5$.
WEEP consistently outperformed baselines in low-noise regimes ($\sigma < 0.5$).
However, performance degrades at higher noise levels ($\sigma > 0.5$), indicating that its hyperparameters are sensitive to noise.
Additionally, we observed that setting $b > 0$ is crucial for preventing vanishing gradients and ensuring convergence of L-BFGS.
\subsection{Real-World Image Denoising}
To demonstrate practical utility, we extend to 2D image denoising using the 'plant' test image with intricate textures.
This image contains real-world noise captured by a Sony A7 II camera \cite{xuRealworldNoisyImage2018}.
The optimization problem is formulated as:
$\minimize_{\bm X} \frac12\|\bm Y - \bm X\|_F^2 + \lambda \sum_{i,j}  \{\phi((\bm D_h \bm X)_{i,j}) + \phi((\bm D_v \bm X)_{i,j})\}$,
where $(\bm{D}_h \bm{X})_{i,j} = X_{i+1,j} - X_{i,j}$ and $(\bm{D}_v \bm{X})_{i,j} = X_{i,j+1} - X_{i,j}$.
We compared WEEP-TV with L2-TV ($\phi\!=\!\|\cdot\|_2^2$), L1-TV and MCP-TV using ADMM with hyperparameters tuned for optimal Structural Similarity Index (SSIM) \cite{wangImageQualityAssessment2004}.

Fig.~\ref{fig:image_denoising_results_with_insets} presents the denoising results. WEEP-TV ($a = 200$, $b = 0$) outperforms all competing methods in PSNR, SSIM, and Feature Similarity Index (FSIM) \cite{zhangFSIMFeatureSimilarity2011}, indicating superior preservation of structural details. Visually, WEEP-TV effectively suppresses noise while maintaining texture sharpness, avoiding the over-smoothing of L2-TV and the staircasing artifacts of L1-TV and MCP-TV. The line profiles further support this result, where WEEP-TV most accurately tracks the ground truth intensity variations.

\section{Conclusion}
We introduced WEEP, a differentiable sparse regularizer based on weakly-convex envelopes. Combining unbiasedness and tunable sparsity strength with a closed-form proximal operator, WEEP demonstrates superior efficacy in denoising and compressed sensing.

Future work includes extending WEEP to structured sparsity (e.g., group sparsity, low-rankness), automated hyperparameter tuning, and deep learning applications, where its smoothness enables end-to-end training.


\bibliography{refs}
\bibliographystyle{modified_IEEEbib}

\end{document}